\documentclass[letterpaper, 10 pt, conference]{ieeeconf}
\IEEEoverridecommandlockouts                              
\let\labelindent\relax
\usepackage{amsfonts}       

\usepackage{amsthm}
\usepackage{mathtools}      
\usepackage{amssymb}        
\usepackage{filecontents}
\usepackage{graphicx}       
\usepackage{marginnote}     
\usepackage{marvosym}       
\usepackage{overpic}        
\usepackage{tabularx}
\usepackage{cite}
\usepackage{color}
\usepackage[linesnumbered,algoruled,boxed,lined,noend]{algorithm2e}
\usepackage[normalem]{ulem}
\usepackage{enumitem}
\usepackage[normalem]{ulem}

\usepackage{lipsum}

\usepackage{comment}
\usepackage[export]{adjustbox}
\usepackage{epstopdf}

\usepackage[font=small]{caption}
\newif\ifdraft
\draftfalse

\ifdraft
\usepackage[paperheight=11in,paperwidth=9.5in,
			left=1.25in,right=1.25in,
			top=0.75in,bottom=0.75in,
			heightrounded,marginparwidth=1.2in,
			marginparsep=0.05in]{geometry}
\usepackage{xcolor}
\usepackage{xargs} 
\usepackage[textsize=footnotesize]{todonotes}
\newcommandx{\nt}[2][1=]{\todo[linecolor=red,
			backgroundcolor=red!10,bordercolor=red,#1]{#2}}
\newcommandx{\jy}[2][1=]{\todo[linecolor=green,
			backgroundcolor=green!10,bordercolor=green,#1]{JY:#2}}
\newcommandx{\kg}[2][1=]{\todo[linecolor=green,
			backgroundcolor=green!10,bordercolor=blue,#1]{KG:#2}}
\newcommandx{\sw}[2][1=]{\todo[linecolor=blue,
			backgroundcolor=orange!10,bordercolor=orange,#1]{SW:#2}}
\else
\usepackage{xcolor}
\usepackage{xargs} 
\usepackage[textsize=footnotesize]{todonotes}
\newcommandx{\nt}[2][1=]{\todo[linecolor=red,
			backgroundcolor=red!10,bordercolor=red,#1]{#2}}
\newcommandx{\jy}[2][1=]{\todo[linecolor=green,
			backgroundcolor=green!10,bordercolor=green,#1]{JY:#2}}
\newcommandx{\kg}[2][1=]{\todo[linecolor=green,
			backgroundcolor=green!10,bordercolor=blue,#1]{KG:#2}}
\newcommandx{\sw}[2][1=]{\todo[linecolor=blue,
			backgroundcolor=orange!10,bordercolor=orange,#1]{SW:#2}}
\fi

\newif\iftwocolumn
\twocolumntrue

\newtheorem{problem}{Problem}
\newtheorem{proposition}{Proposition}[section]

\theoremstyle{definition}

\theoremstyle{remark}

\SetKwProg{Fn}{Function}{}{}
\SetKwComment{Comment}{$\triangleright$\ }{}

\makeatletter
\def\subsubsection{\@startsection{subsubsection}
                                 {3}
                                 {\z@ \hspace*{1mm}}
                                 {0ex plus 0.1ex minus 0.1ex}
                                 {0ex}
                                 {\normalfont\normalsize\itshape}}
\makeatother

\def\toro{\texttt{TORO}\xspace}

\font\titlefont=ptmb at 14.8pt
\title{\titlefont
Toward Efficient Task Planning for Dual-Arm Tabletop Object Rearrangement}
\author{Kai Gao \qquad Jingjin Yu
\thanks{
K. Gao and J. Yu are with the Department of Computer Science, 
Rutgers, the State University of New Jersey, Piscataway, NJ, 
USA. E-Mails: \{{\tt kai.gao, jingjin.yu}\}\hspace*{.25em}
\MVAt \hspace*{.25em}rutgers.edu. This work is supported by NSF awards IIS-1845888 and IIS-2132972.
}%
}

\begin{document}

\maketitle
\thispagestyle{empty}
\pagestyle{empty}


\begin{abstract}
We investigate the problem of coordinating two robot arms to solve 
non-monotone tabletop multi-object rearrangement tasks.
In a non-monotone rearrangement task, complex object-object dependencies exist
that require moving some objects multiple times to solve an instance. 
In working with two arms in a large workspace, some objects must be 
handed off between the robots, which further complicates the planning 
process.
For the challenging dual-arm tabletop rearrangement problem, we develop 
effective task planning algorithms for scheduling the pick-n-place sequence 
that can be properly distributed between the two arms. 
We show that, even without using a sophisticated motion planner, our 
method achieves significant time savings in comparison to greedy approaches 
and naive parallelization of single-robot plans. 
\end{abstract}

\section{Introduction}\label{sec:intro}
\begin{figure}[t]
\vspace{2mm}
    \centering
    \includegraphics[width=0.48\textwidth]{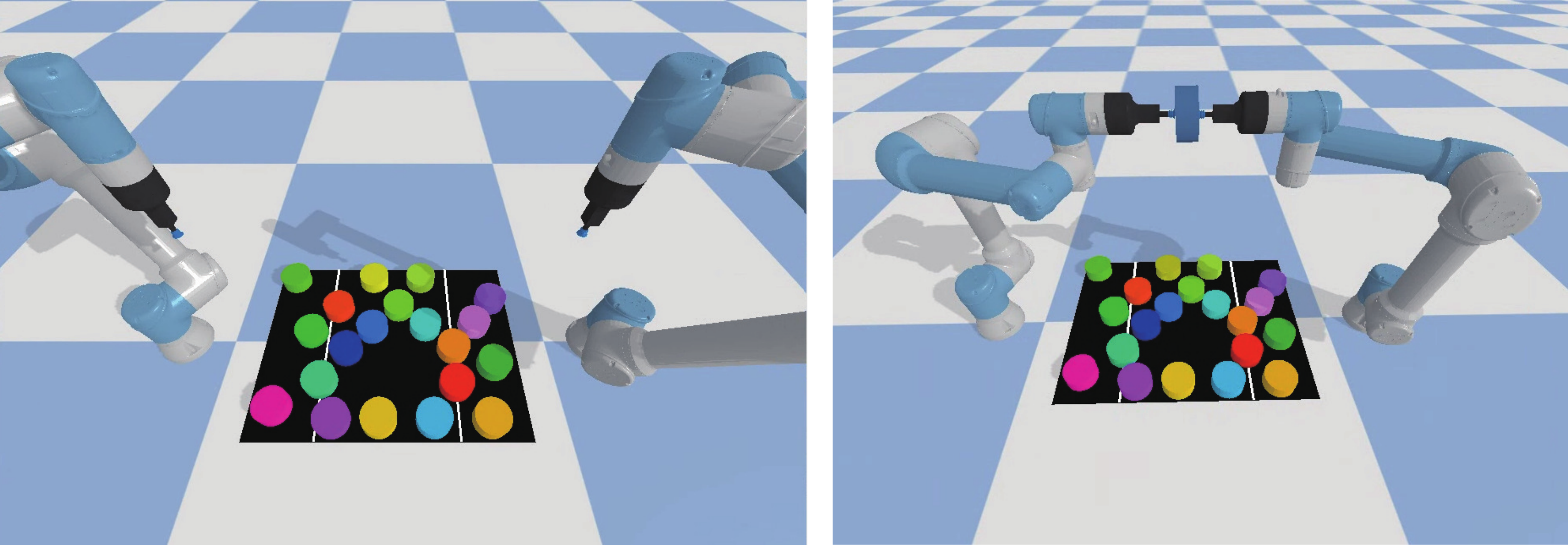}
    \caption{[Left] PyBullet setup for the Cooperative Multi-Robot Rearrangement problem, where only a portion of the environment (the region between two white lines) is reachable by both arms. [Right] Handoff operation at a pre-computed pose above the environment.}
    \label{fig:problem}
\end{figure}

%
%
%
%
In solving multi-object rearrangement problems, employing multiple arms is 
a straightforward way to expand the workspace \cite{shome2020synchronized} 
and increase overall system throughput\cite{shome2021fast}.
%
%
Toward enabling effective dual-arm (and multi-arm) coordination, a difficult 
challenge at present, we investigate the \emph{cooperative dual-arm rearrangement} 
(CDR) problem, where two robot arms' workspaces partially overlap and
each robot is responsible for a portion of the workspace (Fig.~\ref{fig:problem}[Left]). 
In CDR, robot-robot collaboration must be considered for realizing higher 
throughput. 
Certain settings benefit from having more robots, e.g., one arm may hold 
an object while other arms make the goal pose  of the object available. 
This will lead to efficiency gain as compared with using a single robot.
There are also added challenges, however, e.g., some objects must be handed off from one 
robot to another which requires careful synchronization. 

Rearrangement using a single robot is already challenging to optimally solve due 
to complex object-object dependencies \cite{han2018complexity}.
CDR is more involved with an additional arm coordination element. 
%
%
To tackle CDR and minimize task completion makespan, we adopt a lazy buffer
allocation approach \cite{gao2021fast} and solve a rearrangement task in
two phases. 
%
In the first phase, we compute a sequence or schedule of ``primitive rearrangement actions'' without carefully checking feasibility using heuristic search algorithms. 
In particular, it may be infeasible to relocate certain object as planned due to
insufficient free space in the workspace. 
In the second phase, based on the computed schedule, we then allocate intermediate object poses while 
carefully consider potential object-object collisions. 
The heuristic search algorithms are guided by object dependency graphs \cite{van2009centralized,han2018complexity}, which is pre-constructed based on the inherent combinatorial constraints induced by start and goal object arrangements.
With the dependency graph, there is no additional collision checking needed during the scheduling of primitive actions.


This study brings forth several contributions. 
Our method for solving CDR, employing lazy or delayed verification, can 
efficiently compute high-quality solutions for non-monotone instances in a
cluttered environment. 
As a key component of our algorithmic solution, we propose dependency graph 
guided heuristic search procedure for coordinating robot-robot object handoffs 
and individual pick-n-places that supports multiple makespan evaluation 
metrics.
Extensive experiments are conducted to evaluate the performance of proposed 
algorithms and the quality of CDR plans in PyBullet environment with different 
overlap ratio of robot reachable regions, which, in addition to demonstrating 
the effectiveness of our method, provides insights for dual-arm manipulation 
system design.

\textbf{Paper Organization}. The rest of the paper is organized as follows. 
We provide an overview of related literature in Sec.~\ref{sec:related}.
In Sec.~\ref{sec:problem}, we formally define CDR problems, 
discuss the inherent object dependencies and temporary displacements (buffers). 
Then we describe our proposed heuristic search algorithms guided by the pre-computed dependency graph in Sec.~\ref{sec:algorithms}. The algorithms efficiently compute task schedules minimizing makespan under two different metrics. 
In Sec~\ref{sec:buffer}, we outline the lazy buffer allocation approach.
Evaluation follows in Sec.~\ref{sec:evaluation}. 
We conclude with Sec.~\ref{sec:conclusion}.

\section{Related Work}\label{sec:related}
%

{\bf Multi-Object Rearrangement:} Single arm object rearrangement lies within the broader area of Task and Motion Planning (TAMP).
Typical problems in this domain \cite{cosgun2011push, wang2021uniform, gaorunning, wang2021efficient, gao2021fast} involve arranging multiple objects to specific goal poses. Certain variations, however, such as NAMO (Navigation among Movable Obstacles) \cite{stilman2005navigation,  stilman2008planning}, and retrieval problems \cite{ nam2019planning, ZhangLu-RSS-21, vieira2022persistent, huang2022self}, 
focus on clearing out a path for a target object or robot. During this process, they are identifying obstacles that need to be displaced.
Rearrangement may be approached either via simple but inaccurate non-prehensile actions, e.g., pushes \cite{cosgun2011push, king2017unobservable, huang2021dipn, vieira2022persistent}, or more purposeful prehensile actions, such as grasps \cite{krontiris2015dealing, krontiris2016efficiently, wang2021efficient}.

Focusing on inherent combinatorial challenges associated with rearrangement tasks, some planners assume external space for temporary object storage \cite{bereg2006lifting,han2018complexity, nam2019planning,gaorunning}, while others exploit problem linearity to simplify the search  \cite{okada2004environment, stilman2005navigation, stilman2008planning, levihn2013hierarchical}.
By linking multi-object rearrangement to established graph-based problems, efficient algorithms have been obtained for various tasks and objectives  \cite{han2018complexity,gaorunning,bereg2006lifting}.
When external free space is not available, the robot arm needs to allocate collision-free locations for temporary object displacements\cite{krontiris2016efficiently,cheong2020relocate,gao2021fast}.
In our work, we adopt the idea of ``lazy buffer allocation''\cite{gao2021fast} to the dual-arm scenario.

Multi-robot rearrangement requires additional computation on task assignment and coordination.
Ahn et al.\cite{ahn2021coordination} collaborate robots by maximizing the number of ``turn-taking'' moments when one arm is picking an obstacle from the workspace while the other arm is placing the previous obstacle at the external space.
Shome et al.\cite{shome2021fast} assume robot arms pick-n-places simultaneously and only solve monotone rearrangement problems, where each object can move directly to the goal pose.
For objects that need to be moved to a pose outside the reachable region of a robot arm,
the task can be accomplished by coordinating multiple arms to handoff the objects around\cite{shome2020synchronized}.
Cooperative pick-n-place, a related problem to multi-arm rearrangement, is well-studied in the area of printed circuit board assembly tasks\cite{li2008enhancing,moghaddam2016parallelism}. 
These problems do not have ordering constraints on pick-n-place tasks, i.e. all the items can move directly to goal poses in any ordering.
Research in these topics tends to consider multiple grippers equipped on a single arm rather than multiple arms, which reduces system flexibility.
In our work, we compute dual-arm rearrangement plans for dense non-monotone instances without the aid of an external space for temporary storage.

{\bf Dependency Graph:} Dependencies between objects in rearrangement tasks can be naturally represented as a dependency graph, which was first applied to general multi-body planning problems \cite{van2009centralized} and then rearrangement \cite{krontiris2015dealing, krontiris2016efficiently}. 
In rearrangement, different choices of grasp poses and paths give rise to multiple dependency graphs for the same problem instance, which limits the scalability in computing a solution via such representations. 
Prior work \cite{han2018complexity} has applied full dependency graphs to address \toro (Tabletop Object Rearrangement with Overhand Grasps), showing that the problem embeds a NP-hard Feedback Vertex Set problem \cite{karp1972reducibility} and a Traveling Salesperson Problem \cite{papadimitriou1977euclidean}.

More recently, some of the authors \cite{gaorunning} examined another optimization objective, running buffers, which is the size of the external space needed for temporary object displacements in the rearrangement task, and also examined an unlabeled setting,
where goal poses of objects are interchangeable. 
Similar graph structures are also used in other robotics problems, such as packing problems \cite{wang2020robot}. 
Deep neural networks have been also applied to detect the embedded dependency graph of objects in a cluttered environment to determine the ordering of object retrieval \cite{ZhangLu-RSS-21}.

\section{Preliminaries}\label{sec:problem}
\subsection{Problem Statement}
Consider a 2D bounded workspace $\mathcal W \subset \mathbb R^2$ containing a set of $n$ cylindrical objects $\mathcal O=\{o_1, ..., o_n\}$. 
An arrangement $\mathcal A$ of these objects is a set of poses $\{p_1, ..., p_n\}$ with each pose  $p_i=(x_i,y_i)\in \mathcal W$.
$\mathcal A$ is \emph{feasible} if the footprints of objects are inside $\mathcal W$ and pair-wise disjoint.
Outside the workspace, two robot arms $\mathcal R=\{r_1, r_2\}$ are tasked to manipulate objects from a feasible start arrangement $\mathcal A_s$ to another feasible goal arrangement $\mathcal A_g$ (e.g., Fig.~\ref{fig:problem}[Left]). 
Each robot arm $r_i$ has a connected reachable region $\mathcal S(r_i)\subseteq \mathcal W$. 
Robot $r_i$ can only manipulate within $\mathcal S(r_i)$.

\begin{figure}[ht]
    \centering
    \includegraphics[width=0.18\textwidth]{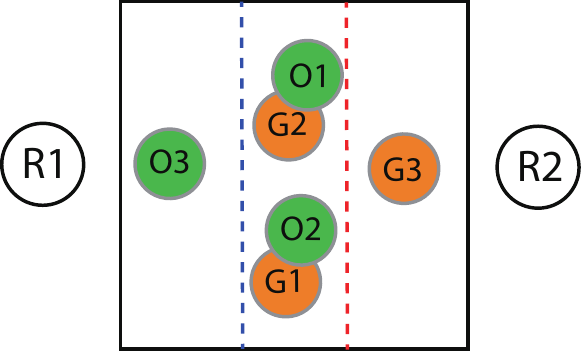}
    \hspace{5mm}
    \includegraphics[width=0.23\textwidth]{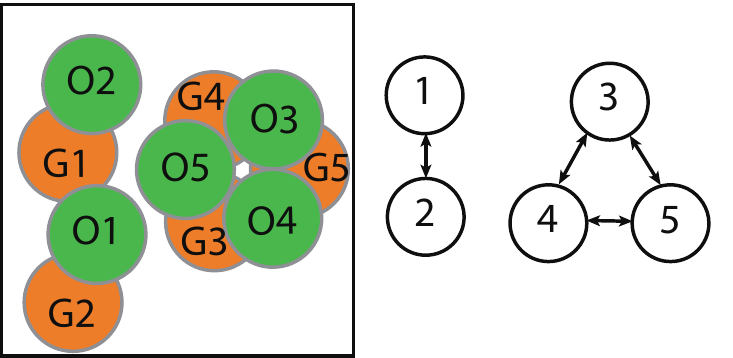}
    \caption{ [Left] A working example of CDR instance with $\rho=0.3$. For each object $o_i$, the start and goal poses are represented by green disc $O_i$ and orange disc $G_i$ respectively. [Right] A CDRF instance and its corresponding dependency graph.}
    \label{fig:CDR-Exp}
\end{figure}

It is assumed that $\mathcal S(r_1)\cup \mathcal S(r_2)=\mathcal W$.
The overlap ratio $\rho$ is defined as $\rho = \dfrac{|\mathcal S(r_1)\cap \mathcal S(r_2)|}{|\mathcal{W}|}$, 
which is the proportion of the environment that can be reached by both robots. 
In Fig.~\ref{fig:CDR-Exp}[Left], we show a workspace with $\rho=0.3$. 
Robot arm $r_1$ can reach objects whose centers are on the left of the red dashed line, which include all the objects at start poses($O_1, O_2, O_3$) and $o_1$, $o_2$ at goal poses($G_1, G_2$). Robot arm $r_2$ can reach objects whose centers are on the right of the blue dashed line, which include all the objects at goal poses($G_1, G_2, G_3$) and $o_1$, $o_2$ at start poses($O_1, O_2$). 
Objects with centers between the dashed lines can be reached by both.

We consider two manipulation primitives: (overhand) \emph{pick-n-place} and \emph{handoff}.
In each pick-n-place, an arm moves above an object $o_j$, grasps it from $\mathcal W$, transfers it atop the workspace, and places it at a collision-free pose  inside $\mathcal W$.
We allow a \emph{handoff} operation when an object needs to cross between  $\mathcal S(r_1)\backslash \mathcal S(r_2)$ and $\mathcal S(r_2)\backslash \mathcal S(r_1)$.
In each handoff, an arm, say $r_1$, grasps an object $o_j$ from $\mathcal W$, 
passes the object to the other arm $r_2$ at the predefined handoff pose above the environment.
$r_2$ receives $o_j$ and places it at a collision-free pose  in $\mathcal W$.
A handoff is shown in Fig.~\ref{fig:problem}[Right].


As for the motion of each arm $r_i$, $\mathcal C^i_{free}$ is a set of collision-free configurations for $r_i$ without considering collisions with workspace objects or the other robot arm.
Throughout the rearrangement, each arm has $n+1$ motion modes: transit mode $TS$ (moving on its own) and transfer modes $TF_j$ (transferring the object $o_j$), $\forall j=1,...,n$.
Therefore, the state space $\Gamma_i$ of an arm $r_i$ is the Cartesian product $\mathcal C^i_{free} \times M$, where $M$ is the motion mode set. 
A path of a robot arm $r_i$ is defined as $\pi_i:[0,T]\rightarrow \Gamma_i$ with $\pi_i[0]=\pi_i[T]=(q^*_i, TS)$, where $T$ is the makespan of the rearrangement plan and $q^*_i$ is the rest pose of arm $r_i$.

The rearrangement problem we study seeks a path set $\Pi=\{\pi_1, \pi_2\}$, such that robot paths are collision-free: for all $t\in [0,T]$, arm $r_1$ at state $\pi_1[t]$ does not collide with arm $r_2$ at state $\pi_2[t]$.
Additionally, the solution quality is evaluated by the makespan under two levels of fidelity:
\begin{enumerate}
    \item (MC) manipulation cost based makespan;
    \item (FC) estimated execution time based makespan.
\end{enumerate}
In MC (manipulation cost) makespan, we assume the manipulation time dominates 
the execution time, which is often the case in today's systems. MC makespan is defined as the number of \emph{steps} for the rearrangement task. In each step, 
each robot arm can complete an individual pick-n-place, execute a coordinated handoff, or idle.

In FC (full cost) makespan, solution quality is measured by the estimated 
complete time based on four parameters: 
maximum horizontal speed $s$ of the end-effectors, the execution time of each pick ($t_g$), place ($t_r$), and handoff ($t_h$). The details of FC metric and its corresponding heuristic search method is presented in the appendix (Sec.\ref{sec:CDR-KC})

Given the setup, the problem studied in this paper can be summarized as:

\begin{problem}[CDR: Collaborative Dual-Arm Rearrangement]
Given feasible arrangements $\mathcal A_s$, $\mathcal A_g$, and makespan metric (MC or FC), determine a collision-free path set $\Pi$ for $\mathcal R$ moving objects from $\mathcal A_s$ to $\mathcal A_g$ minimizing the makespan $T$.
\end{problem}

We also investigate two special cases of CDR.
On one hand, when $\rho=1$, $\mathcal S(r_1)=\mathcal S(r_2)=\mathcal W$, 
which yields to Collaborative Multi-Arm Rearrangement with Full Overlap (CDRF). 
In this case, handoff is not needed since each robot can execute pick-n-place operations individually inside the workspace. 
On the other hand, when $\rho=0$, $\mathcal S(r_1)\bigsqcup \mathcal S(r_2)=\mathcal W$, 
which yields to Collaborative Multi-Arm Rearrangement with no Overlap (CDRN). 



\subsection{Dependency Graph and Object Displacement}
In multi-object rearrangement, minimizing the number of pick-n-places directly leads to increased throughput.
Previous works\cite{han2018complexity}\cite{gaorunning} treat the combinatorial challenge in the single robot rearrangement problem with a dependency graph structure $\mathcal G$.
$o_i$ \emph{depends on} $o_j$ when $o_i$'s goal pose overlaps with $o_j$'s start pose.
In this case, $o_j$ needs to be moved away before $o_i$ is moved to the goal.
$\mathcal G$ is constructed with vertices representing objects and arcs representing the dependencies.

If $\mathcal G$ is acyclic, then the rearrangement problem is \emph{monotone}, where objects can be directly moved to goal poses based on the topological ordering implied by $\mathcal G$.
On the other hand, if there are cycles in the dependency graph, then the 
problem is \emph{non-monotone}; some object(s) in the cycles need to be 
displaced to ``break'' the cycles. For a single arm, the displaced object 
must be placed down at some temporary pose. 
In a dual-arm system, the extra arm can temporarily hold an object to be displaced.
For example, Fig.~\ref{fig:CDR-Exp}[Right] shows a CDRF instance and its corresponding dependency graph $\mathcal G$.
In this case, the cycle of $o_1$ and $o_2$ can be broken by letting $r_1$ move $o_1$ and $r_2$ move $o_2$ at the same time to the goal poses.
However, the cycle among object $o_3$, $o_4$ and $o_5$ cannot be broken without an extra action moving an object (e.g. $o_3$) to some temporary pose.

The free space for a temporary object displacement is called a \emph{buffer}; the planner must allocate buffers inside the workspace.
We apply a lazy buffer allocation \cite{gao2021fast} which is shown to be efficient in computing high quality rearrangement plans for a single arm.
In the framework, we first compute a valid schedule of primitive actions, which only indicates whether the objects are handed off, moved to goal poses, or to buffers.
And then based on the primitive plan, we allocate proper buffer locations as needed. The scheduling problem of primitive actions is discussed in Sec.~\ref{sec:algorithms}, and the buffer allocation process is discussed in Sec.~\ref{sec:buffer}.



\section{Task Scheduling for CDR}\label{sec:algorithms}

In this section, we study the scheduling problem of primitive actions in CDR. 
A primitive action can be represented by a tuple $(r_i, o_j,v )$ indicating that $r_i$ is tasked to transfer $o_j$ to new status $v$, where $v$ is one kind of the object status in $\{S,G,H,\mathcal B(r_1),\mathcal B(r_2)\}$.
$S$, $G$, $H$, $\mathcal B(r_1)$, and $\mathcal B(r_2)$ represent that the object is at the start pose, goal pose, handoff pose, a buffer in $\mathcal S(r_1)$, and a buffer in $\mathcal S(r_2)$ respectively.
For each primitive action $a$, the schedule provides the starting time and estimated ending time of $a$.
Noting that buffer locations are not determined in the task scheduling process, 
they are categorized into $\mathcal B(r_1)$ and $\mathcal B(r_2)$ to indicate the arm that they are reachable by.
The buffer allocation is discussed in Sec.~\ref{sec:buffer}.
Since the scheduling problem only depends on the availability of goal poses and reachability of robot arms to the start and goal poses, 
the constraints can be fully expressed by the dependency graph.
We describe a dependency graph guided heuristic search method for CDR under MC makespan metrics (Sec.~\ref{sec:CDR-MC}). Its variant for FC makespan is shown in the appendix (Sec.~\ref{sec:CDR-KC}). 

\subsection{Arrangement Space Heuristic Search for CDR in MC Makespan (MCHS)}\label{sec:CDR-MC}
When the makespan is counted in MC metric, 
we can assume primitive actions are executed simultaneously in the task scheduling process without loss of generality and the makespan is counted as the number of \emph{action steps} in the schedule.
Each action at action step $t$ is scheduled to start at time $t$ and end at time $t+1$.
Similar to the single arm rearrangement problem, we search for a plan by maintaining a search tree in the arrangement state space.
Each arrangement state in the tree represents an object arrangement, which can be expressed as a mapping $\mathcal L: \mathcal O\rightarrow \{S,G,\mathcal B(r_1),\mathcal B(r_2)\}$. The arrangement state represents a moment when all objects are stably located in the workspace and both arms are empty.

Arrangements are connected by primitive actions of two arms, 
which consist of one out of the three possible options: 
(1) individual pick-n-places: one or both arms move objects to poses that are currently collision-free; 
(2) coordinated pick-n-places: move an object to the goal pose while the other arm tries to clear away the last obstacle at its goal pose, e.g. swapping poses of $o_1$ and $o_2$ in Fig.~\ref{fig:CDR-Exp}[Left]; 
(3) a coordinated handoff for an object when its start and goal poses are not both reachable by either arm.
We ignore primitive actions that hold an object for more than one action step due to the unnecessary loss of efficiency.

\begin{figure}
    \centering
    \includegraphics[width=0.5\textwidth]{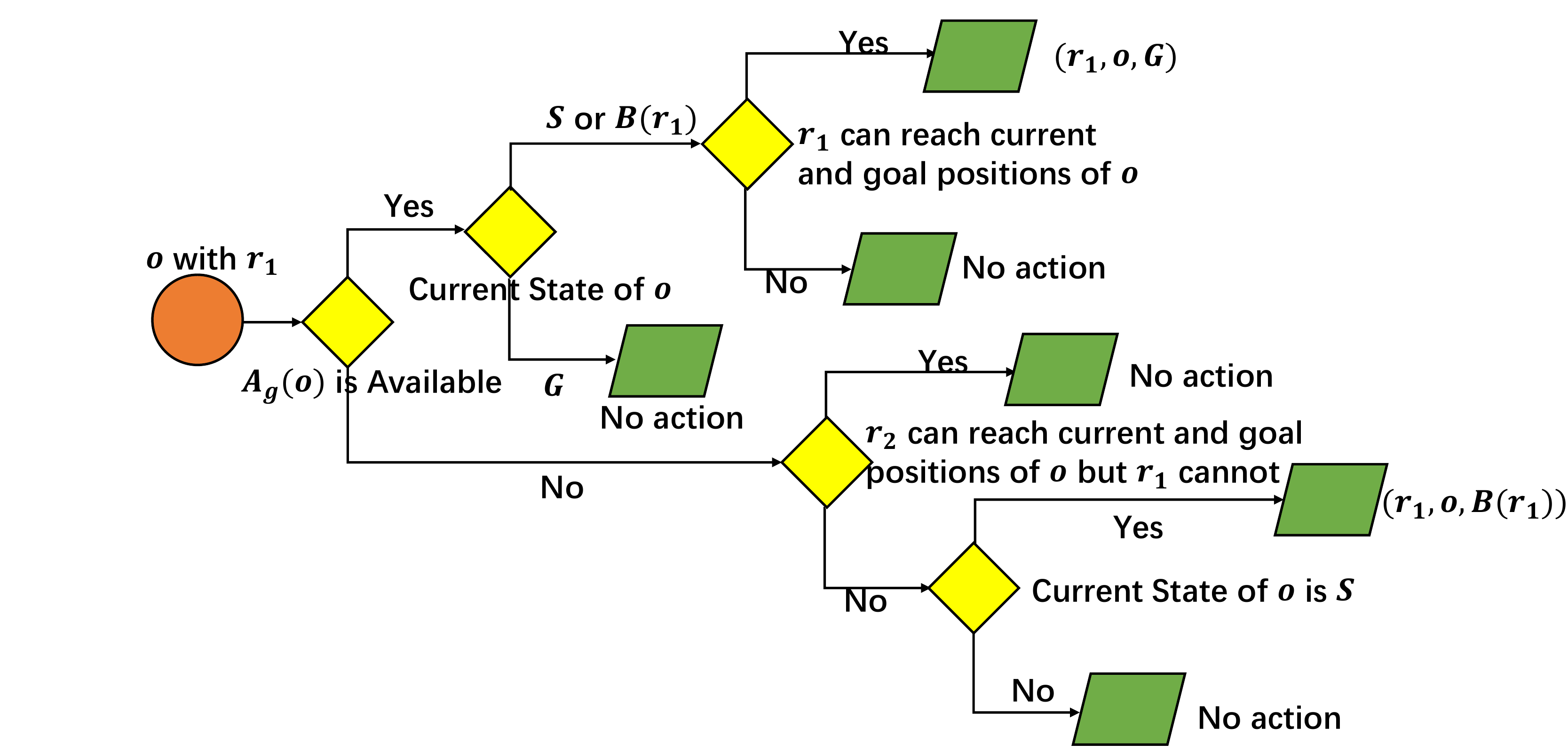}
    \includegraphics[width=0.5\textwidth]{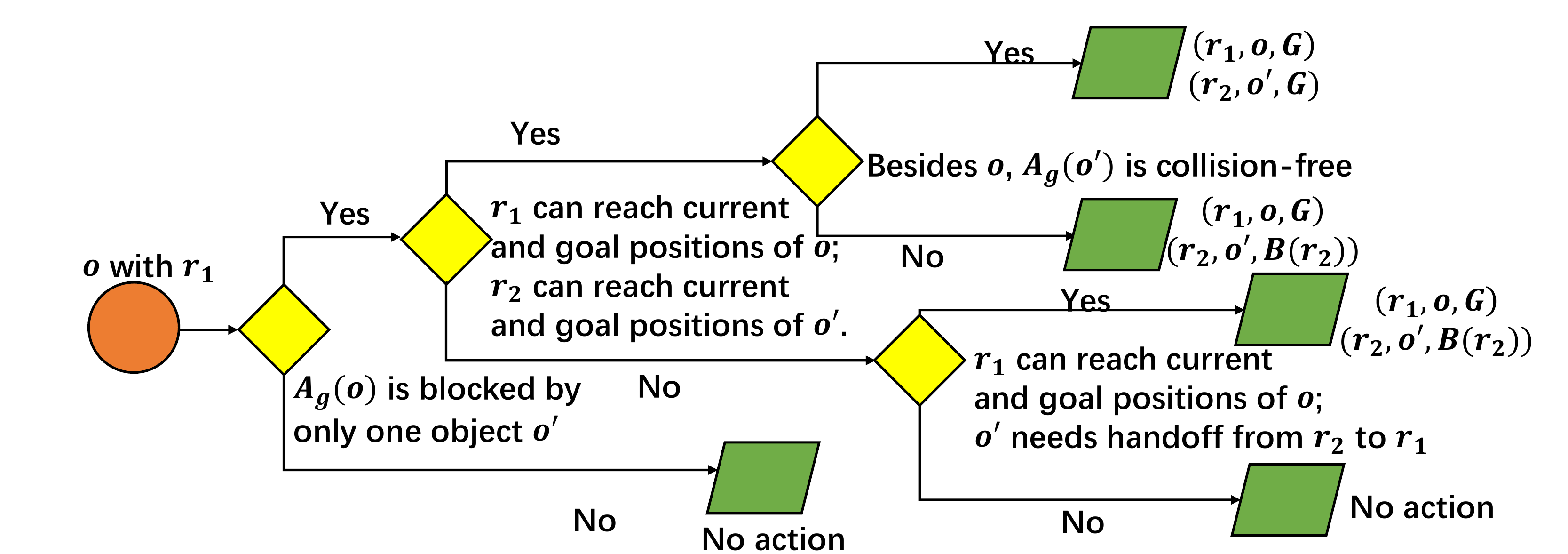}
    \includegraphics[width=0.5\textwidth]{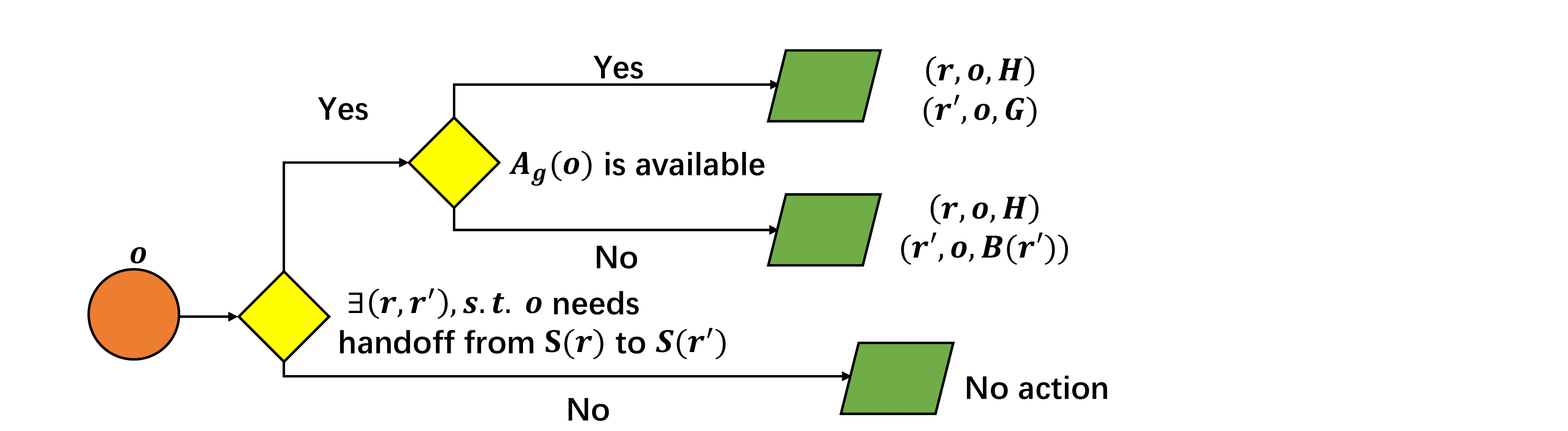}
    \caption{
    Decision process of generating possible primitive actions.
    [Upper] The decision process for $r_1$ to manipulate $o\in \mathcal O$ without cooperation with the other arm. That for $r_2$ is equivalent. 
    [Middle] The decision process for $r_1$ to manipulate an object $o\in \mathcal O$ when only one object $o'$ is blocking the goal pose of $o$. 
    That for $r_2$ is equivalent.
    [Bottom] The decision process when an object $o\in \mathcal O$ needs handoff.
    }
    \label{fig:flow-charts}
\end{figure}

Fig.~\ref{fig:flow-charts} shows the flow charts indicating the decision process of generating possible primitive actions in each arrangement state.
Fig.~\ref{fig:flow-charts}[Upper] shows the decision process for $r_1$ to pick-n-place $o\in \mathcal O$ without cooperation with the other arm. 
That for $r_2$ is equivalent.
Fig.~\ref{fig:flow-charts}[Middle] shows the decision process for $r_1$ to manipulate an object $o\in \mathcal O$ when only one object $o'$ is blocking the goal pose of $o$. In this case, two arms cooperate to send $o$ to the goal.
That for $r_2$ is equivalent.
Fig.~\ref{fig:flow-charts}[Below] shows the decision process when an object $o\in \mathcal O$ needs handoff.
In the process, the availability of goal poses can be checked in the precomputed dependency graph and the reachability of arms is indicated by the object status in the arrangements. Therefore, with the dependency graph, no additional collision-checking is needed in the task scheduling process of CDR.

\begin{figure}
    \centering
    \includegraphics[width=0.25\textwidth]{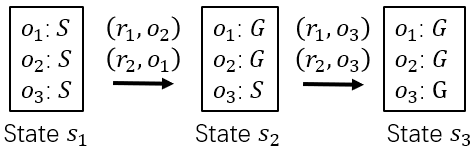}
    \hspace{3mm}
    \includegraphics[width=0.12\textwidth]{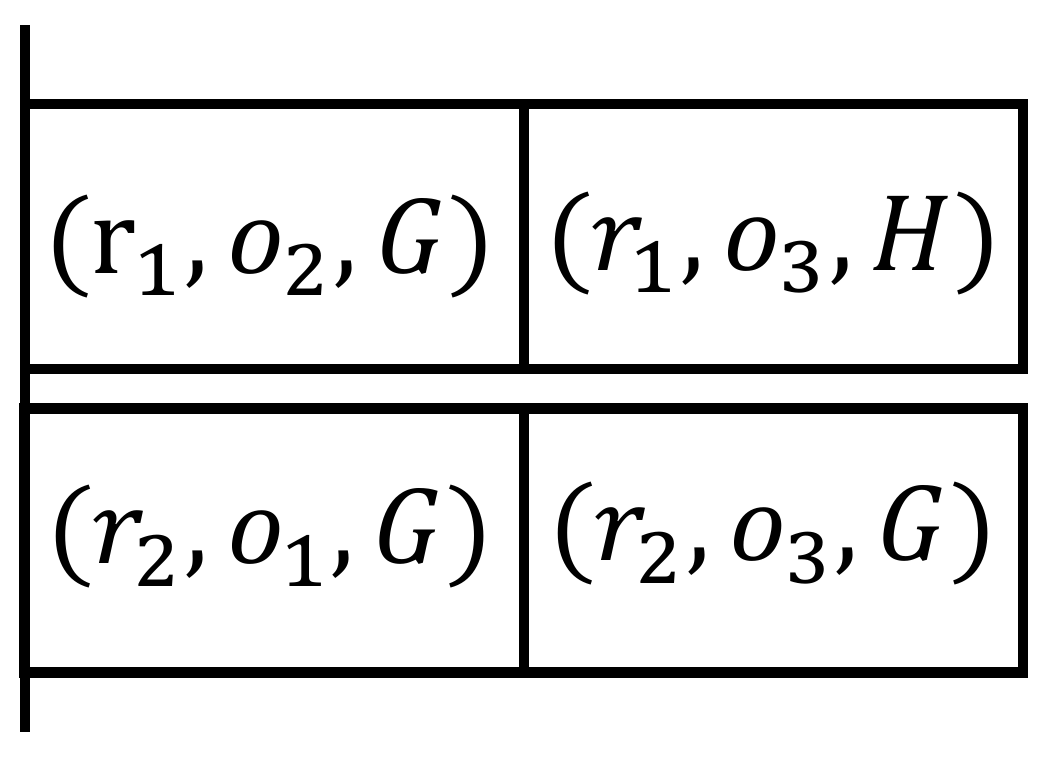}
    \caption{[Left] A path on the arrangement state space for the instance in Fig.~\ref{fig:CDR-Exp}[Left] under MC metric. [Right] The corresponding schedule of the example instance in Fig.~\ref{fig:CDR-Exp}[Left].}
    \label{fig:MC-Plan}
\end{figure}

For the working example in Fig.~\ref{fig:CDR-Exp}[Left], a corresponding rearrangement plan, in the form of a path in the arrangement state space is presented in Fig.~\ref{fig:MC-Plan}[Left]. 
At the start arrangement state $s_1$, $r_1$ and $r_2$ move $o_2$ and $o_1$ respectively to the goal poses, 
which yields the next arrangement state $s_2$.
At $s_2$, two arms coordinate to execute a handoff delivering $o_3$ to the goal and reach the goal arrangement state $s_3$. 
By concatenating primitive actions from $s_1$ to $s_3$, we obtain a corresponding task schedule for the instance (Fig.~\ref{fig:MC-Plan}).

The arrangement state heuristic search based on MC makespan (MCHS) explores the arrangement state space in a best-first manner, always developing the arrangement state $s$ with the lowest $f(s)=g(s)+h(s)$.
The $g$ function is number of action steps from the start arrangement state to the current arrangement state. For example, 
in Fig.~\ref{fig:MC-Plan}, $g(s_3)=2$.
For each state $s$, we define the heuristic function $h(s)$ based on the needed cost to move all objects directly to goals. 

The details are presented in Algo.\ref{alg:MC-heuristic}.
In CDR, costs can be either exclusive cost of $r_1$, or exclusive cost of $r_2$, or shared cost of $r_1$ and $r_2$. 
For example, on one hand, a pick-n-place in region $\mathcal S(r_1)\backslash \mathcal S(r_2)$ can only be executed by $r_1$, 
so it leads to an exclusive cost of $r_1$. 
On the other hand, a pick-n-place in the overlapping region $\mathcal S(r_1)\bigcap \mathcal S(r_2)$ can be executed by either arm so the cost can be shared by both arms.
Based on this observation, we count the costs cost[$r_1$], cost[$r_2$], and sharedCost in the heuristic and initialize them as 0 (Line 1).
We ignore the objects at the goal poses (Line 3).
For the remaining objects, the cost category is determined by $r_1$, $r_2$'s reachability to the current pose (buffer or start pose) and goal pose of the object.
ArmSet is the set of arms able to reach both the current pose and goal pose (Line 4).
If both arms are in armSet (Line 5), then the cost of moving $o_i$ to the goal will be added to sharedCost, which means that it is a pick-n-place in the overlapping region.
If only one arm is in armSet (Line 6-7), then the cost of moving $o_i$ to the goal will be added to the cost of the specific arm, 
which means that this action happens out of reach of the other arm.
If neither arm is in armSet (Line 8-10), then the object needs a handoff to the destination.
In this case, both arms need to take one action for the handoff operation.
Finally, on one hand, if the difference of the exclusive costs is smaller than sharedCost (Line 11), then $h(s)$ is half of the total cost (Line 12). 
In other words, the cost of two arms may be balanced by splitting the sharedCost.
On the other hand, if the difference of the exclusive costs is too large (Line 13), then the unoccupied arm bears all the shared cost, and $h(s)$ equals the cost of the busy arm.

\begin{algorithm}
\begin{small}
    \SetKwInOut{Input}{Input}
    \SetKwInOut{Output}{Output}
    \SetKwComment{Comment}{\% }{}
    \caption{Manipulation Cost Search Heuristic}
		\label{alg:MC-heuristic}
    \SetAlgoLined
		\vspace{0.5mm}
    \Input{$s$: current state}
    \Output{h: heuristic value of $s$}
		\vspace{0.5mm}
		cost[$r_1$], cost[$r_2$], sharedCost $\leftarrow 0,0,0$\\
		\For{$o_i\in \mathcal O$}{
		\lIf{$\mathcal L(o_i)$ is $G$}{continue}
		armSet $\leftarrow$ ReachableArms($s$, $o_i$, $\{\mathcal L(o_i), G\}$)\\
		\lIf{armSet is $\{r_1, r_2\}$}{sharedCost$++$}
		\lElseIf{armSet is $\{r_1\}$}{cost[$r_1$]$++$}
		\lElseIf{armSet is $\{r_2\}$}{cost[$r_2$]$++$}
		\Else{
		cost[$r_1$] $++$;\\
		cost[$r_2$] $++$;
		}
		}
		\vspace{1mm}
		\lIf{$\|$cost[$r_1$]-cost[$r_2$] $\|\leq$ sharedCost}{
		\\ \Return (cost[$r_1$]+cost[$r_2$]+sharedCost)/2
		}
		\lElse{
		\Return max(cost[$r_1$], cost[$r_2$])
		}
\end{small}
\end{algorithm}

Since $cost[r_1]$ and $cost[r_2]$ provide lower bounds of necessary manipulations for each arm, $h(s)$ is a lower bound of the manipulation cost from the state $s$ to the goal arrangement. Therefore, we have:

\begin{proposition}
The manipulation cost based heuristic $h(s)$ is consistent and admissible.
\end{proposition}

\section{Buffer Allocation}\label{sec:buffer}
In this section, we allocate feasible buffer locations based on the primitive plan computed in Sec.~\ref{sec:algorithms} following the idea of lazy buffer allocation\cite{gao2021fast}. 
Given the primitive plan, we know not only the specific timespans that objects stay in buffers, 
but also the staying poses of the other objects in the timespans, which are treated as fixed obstacles that the buffer location needs to avoid.
Besides the obstacles, 
the buffer location of an object moved to $\mathcal B(r)$ in the primitive plan is restricted to be inside $\mathcal S(r)$, the reachable region of $r$.

\begin{figure}
    \centering
    \includegraphics[width=0.2\textwidth]{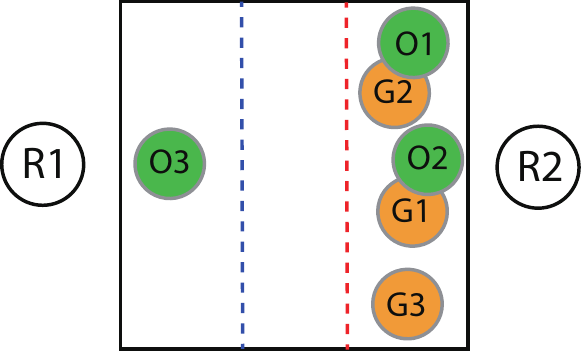}
    \vspace{5mm}\\
    \includegraphics[width=0.49\textwidth]{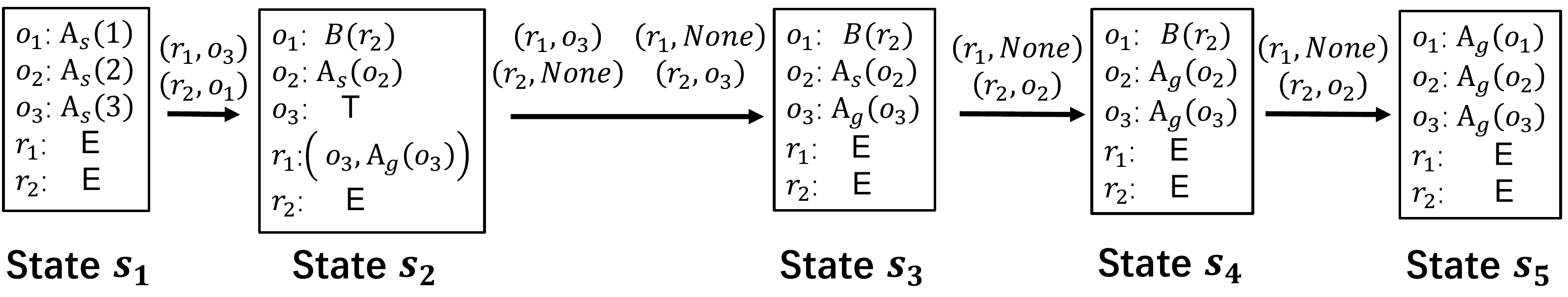}
    \vspace{1mm}
    \caption{[Top] A CDR instance that needs a buffer. [Bottom] An rearrangement plan for the example.}
    \label{fig:CDR-Buffer_Exp}
\end{figure}

We illustrate the buffer allocation process with a CDR instance in Fig.~\ref{fig:CDR-Buffer_Exp}[Top], which is a variant of the example in Fig.~\ref{fig:CDR-Exp}[Left]. 
When $o_1$ and $o_2$ are in $\mathcal S_2 \backslash \mathcal S_1$,
at least one of them needs to be moved to buffer locations to break the cycle in the dependency graph.
A rearrangement plan is shown in Fig.~\ref{fig:CDR-Buffer_Exp}[Bottom]. 
In this plan, $o_1$ is moved to a buffer when $o_2$ is at $\mathcal A_s(o_2)$, and $o_3$ waits for a handoff. 
And $o_1$ leaves the buffer when $o_2$ and $o_3$ are both at goal positions.
Therefore, the buffer location of $o_1$ needs to be collision-free with $o_2$ at $\mathcal A_s(o_2)$, $\mathcal A_g(o_2)$ and $o_3$ at $\mathcal A_g(o_3)$. 
Additionally, it is restricted in $\mathcal S(r_2)$.
When multiple objects stay in buffers at the same time, the buffer locations need to be allocated disjointedly.
In this paper, we search for feasible buffer locations by sampling candidates in the reachable region and check collisions with the fixed obstacles indicated by the primitive plan.
The same as TRLB\cite{gao2021fast}, 
when buffer allocation fails, 
we accept partial plans and conduct a bi-directional search in the arrangement state to connect the start and goal arrangement.

\section{Experimental Studies}\label{sec:evaluation}
To evaluate the effectiveness of our dual-arm task planning algorithms, we 
integrate them with a simple priority-based motion planner. Essentially, 
in the presence of potential arm-arm conflict in the workspace, the motion 
planner will give the arm that is moving first priority while the second arm 
yields.
The yielding arm will either take a detour to the target pose or go back to the rest pose and wait for the execution of the other arm.
We make this choice, instead of using more sophisticated sampling-based
asymptotically optimal planners, to highlight the benefit of our task planner. 
%

\begin{figure}[h]
    \centering
    \includegraphics[width=0.4\textwidth]{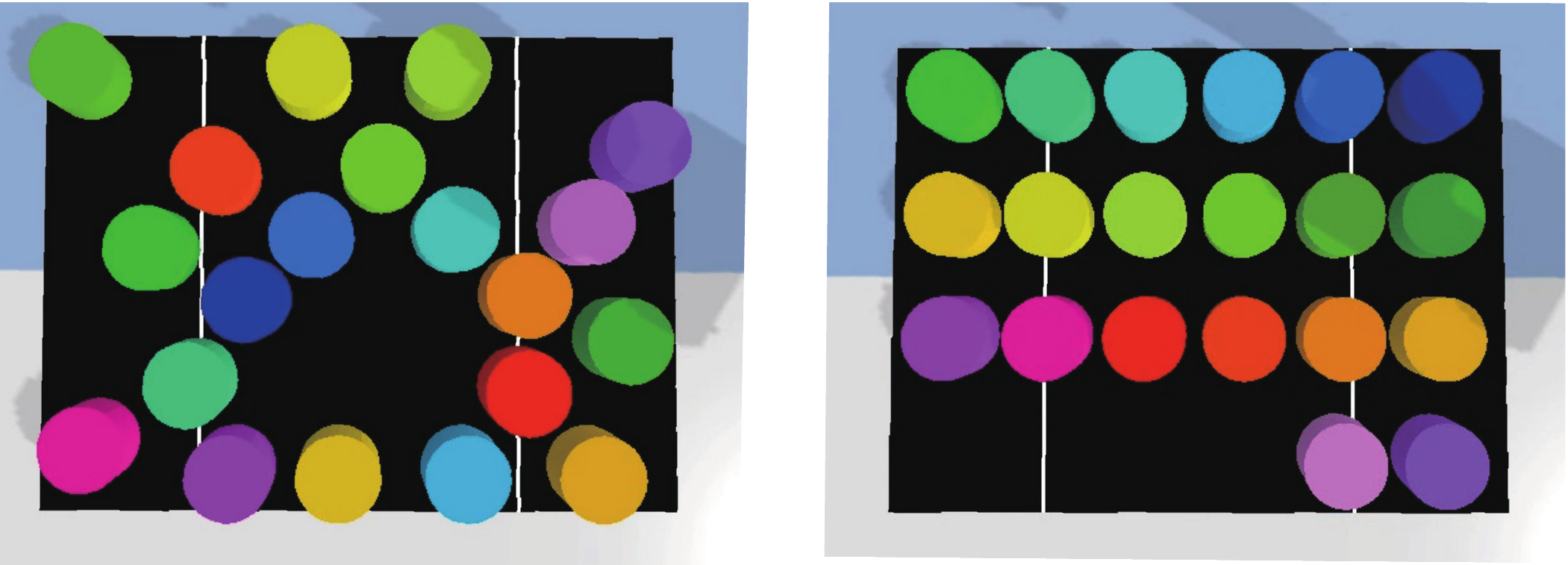}
    \caption{A 20-cylinder example of rearrangement instance with $D=0.4$ and $\rho=0.5$, moving object from a randomly sampled start arrangement [Left], to an organized goal arrangement [Right].}
    \label{fig:organized}
\end{figure}

The proposed algorithms are implemented in Python; we choose objects to be uniform cylinders.
In each instance, robot arms move objects from a randomly sampled start arrangement to an ordered goal arrangement (see, e.g., Fig.~\ref{fig:organized}).
Besides different overlap ratios $\rho$, we test our algorithms in environments with different density levels $D$, defined as the proportion of the tabletop surface occupied by objects, 
i.e., $D:=(\Sigma_{o_i\in \mathcal O} S_{o_i})/S_{\mathcal W}$, 
where $S_{o_i}$ is the base area of $o_i$ and $S_{\mathcal W}$ is the area of $\mathcal W$.
The computed rearrangement plans are executed in PyBullet with UR-5e robot arms.

For the FC makespan, we propose an interval state space heuristic search algorithm (FCHS) in Sec.\ref{sec:CDR-KC}.
In experiments, we let the execution time of a pick $t_g$, a place $t_r$, and a handoff $t_h$ equal to the time spent for an arm traveling across the diameter of the workspace $t_d$, i.e. $t_g=t_r=t_h=t_d$.
This setting mimics general manipulation scenarios in the industry, where a successful pick-n-place relies on accurate pose estimation of the target object, a careful picking to firmly hold the object, 
and a careful placing to stabilize the object at the desired pose.
The experiments are executed on an Intel$^\circledR$ Xeon$^\circledR$ CPU at 3.00GHz. 
Each data point is the average of $20$ test cases except for unfinished trials, if any, given a time limit of $300$ seconds for each test case.

\begin{figure}[h]
    \centering
    \includegraphics[width=0.48\textwidth]{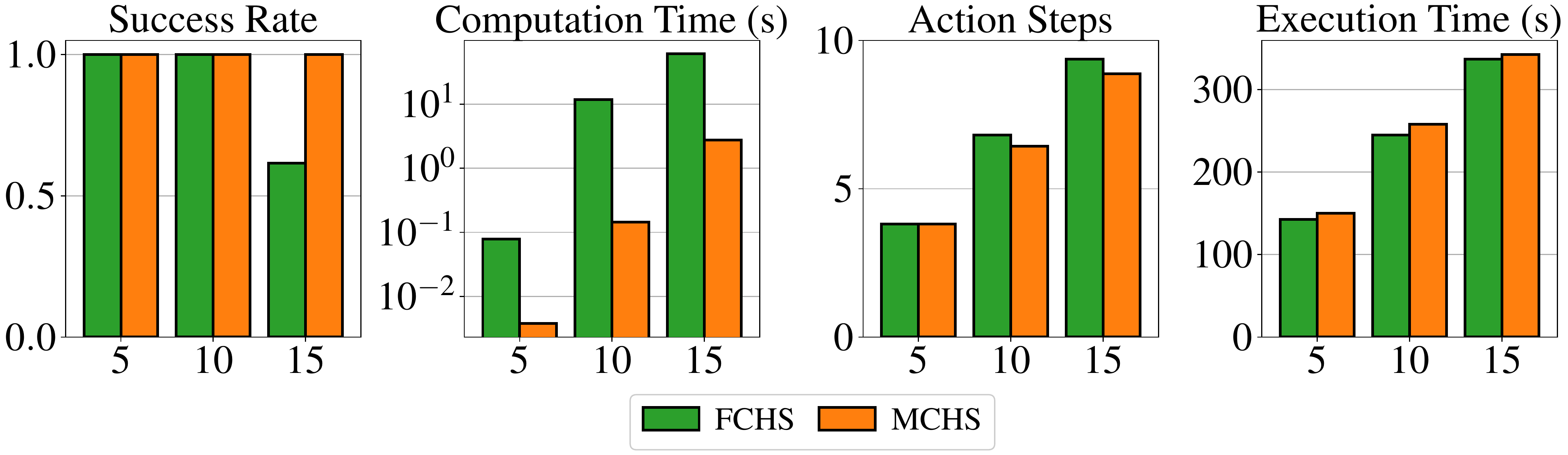}
    \caption{Performance of FCHS and MCHS in instances with $\rho=0.5$, $D=0.3$. The x-axis represents the number of objects.}
    \label{fig:MC_vs_FC}
\end{figure}

\textbf{Manipulation versus Full Cost as Proxies.} In Fig.\ref{fig:MC_vs_FC}, we first compare optimal task schedules in both MC and FC makespans in instances with $\rho=0.5$, $D=0.3$.
While FCHS shows slight advantage in actual execution time, it incurs significantly more computation time and is more prone to planning failure. 
The results suggest that MC is more suitable as a proxy for optimizing CDR planning process.
We also compare the execution time of MC-optimal and FC-optimal plans in instances under different density levels and overlap ratios and observe similar outcome. 
Based on the observation, for later experiments, MCHS is used exclusively.

\textbf{Comparison with Baselines.} We compare the proposed algorithms with two baseline methods:
\begin{enumerate}
    \item \emph{Single-MCHS-Split:} A schedule of primitive actions for a single arm is first computed, minimizing the MC makespan, i.e. the total number of actions.
    Then, the tasks are assigned to the two arms as evenly as possible.
    A handoff is coordinated if the manipulating object is moving between $\mathcal S(r_1)$ and $\mathcal S(r_2)$.
    \item \emph{Greedy}: Each arm prioritizes the manipulations moving objects from buffers to goals.
    When there is no such manipulation available, the arm will choose the object closest to the current end-effector position.
\end{enumerate}

\begin{figure}[h]
    \centering
    \includegraphics[width=0.48\textwidth]{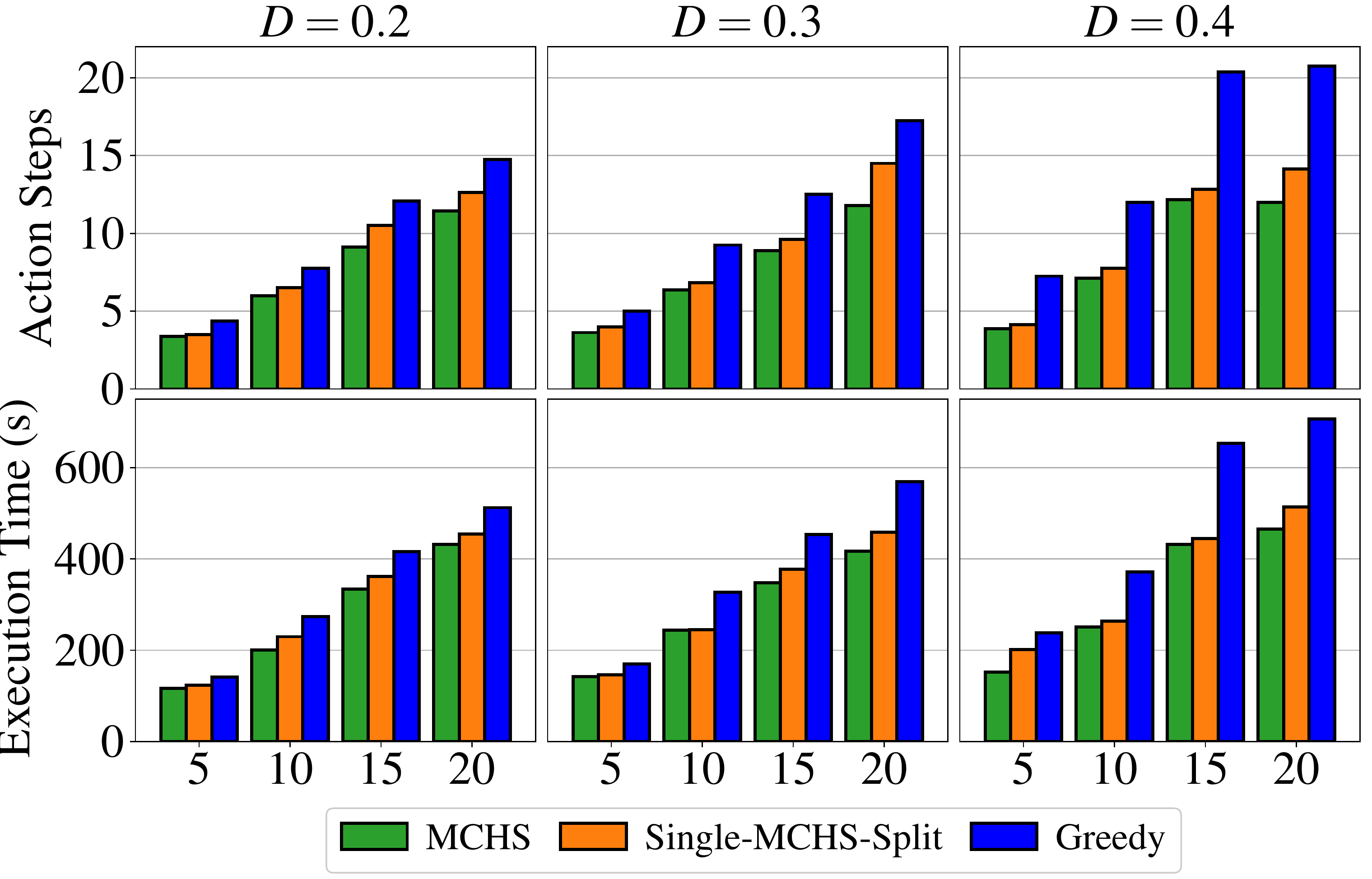}
    \caption{Algorithm comparison in environments with different density levels $D$. The x-axis represents the number of objects.}
    \label{fig:density_results}
\end{figure}

In Fig.~\ref{fig:density_results}, we compare MCHS to the baseline algorithms in the environments with $\rho=0.5$ and different density levels $D$.
Comparing with Single-MCHS-Split, MCHS saves up to $10\%$ execution time, which is fairly significant for logistic applications.
We note that Single-MCHS-Split, guided by dependency graph, already performs quite well in dense environments.
The execution time gap comes from failures in coordinating Single-MCHS-Split plans, e.g. an arm idles for a few action steps waiting for a handoff or an arm holds an object for a few action steps waiting for obstacle clearance.
Comparing with the greedy method, MCHS saves more execution time as $D$ increases. Specifically, in 20-cylinder instances with $D=0.4$, the execution time of MCHS plans is $35\%$ shorter than greedy plans. 
Without a long-term plan in mind, greedy planning incurs more temporary object displacements.

\begin{figure}[h]
    \centering
    \includegraphics[width=0.4\textwidth]{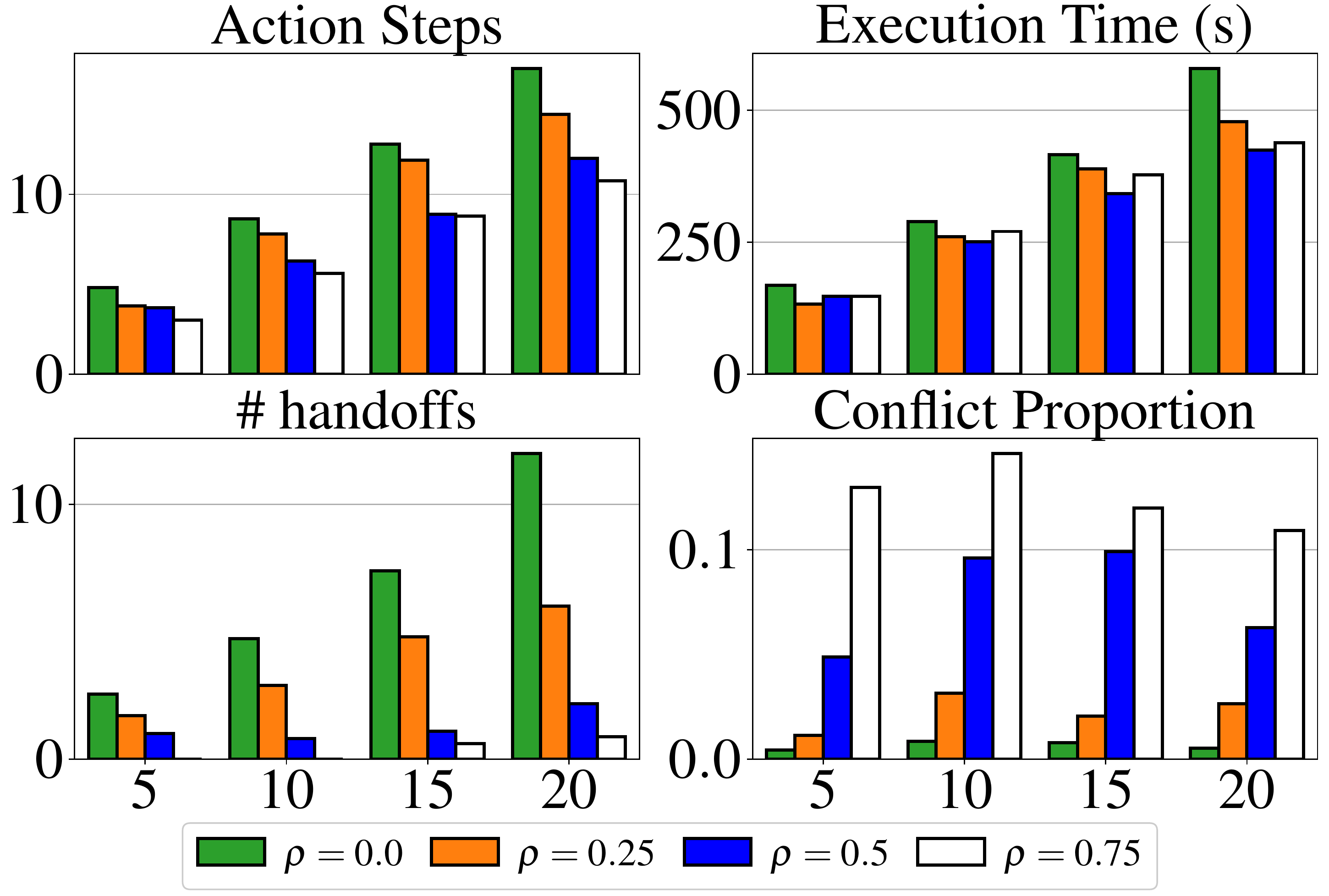}
    \caption{Evaluation of instances with different $\rho$. The x-axis represents the number of objects.}
    \label{fig:various_overlap}
\end{figure}

In instances with $D=0.3$, we also evaluate the tendency of the makespan and execution time as $\rho$ varies.
The conflict proportion shows the proportion of execution time for an arm to yield to the other arm during the execution.
The results are shown in Fig.~\ref{fig:various_overlap}.
As $\rho$ increases, the shared area of robot arms expands. 
On one hand, the number of objects that need handoffs decreases. So does the number of action steps in the schedule.
On the other hand, we see an increase of path conflicts.
In instances with larger $\rho$, robot arms spend more time on yielding.
Based on the two factors, there is a shallow ``U'' shape in execution time as $\rho$ increases.

\begin{figure}[h]
    \centering
    \includegraphics[width=0.3\textwidth]{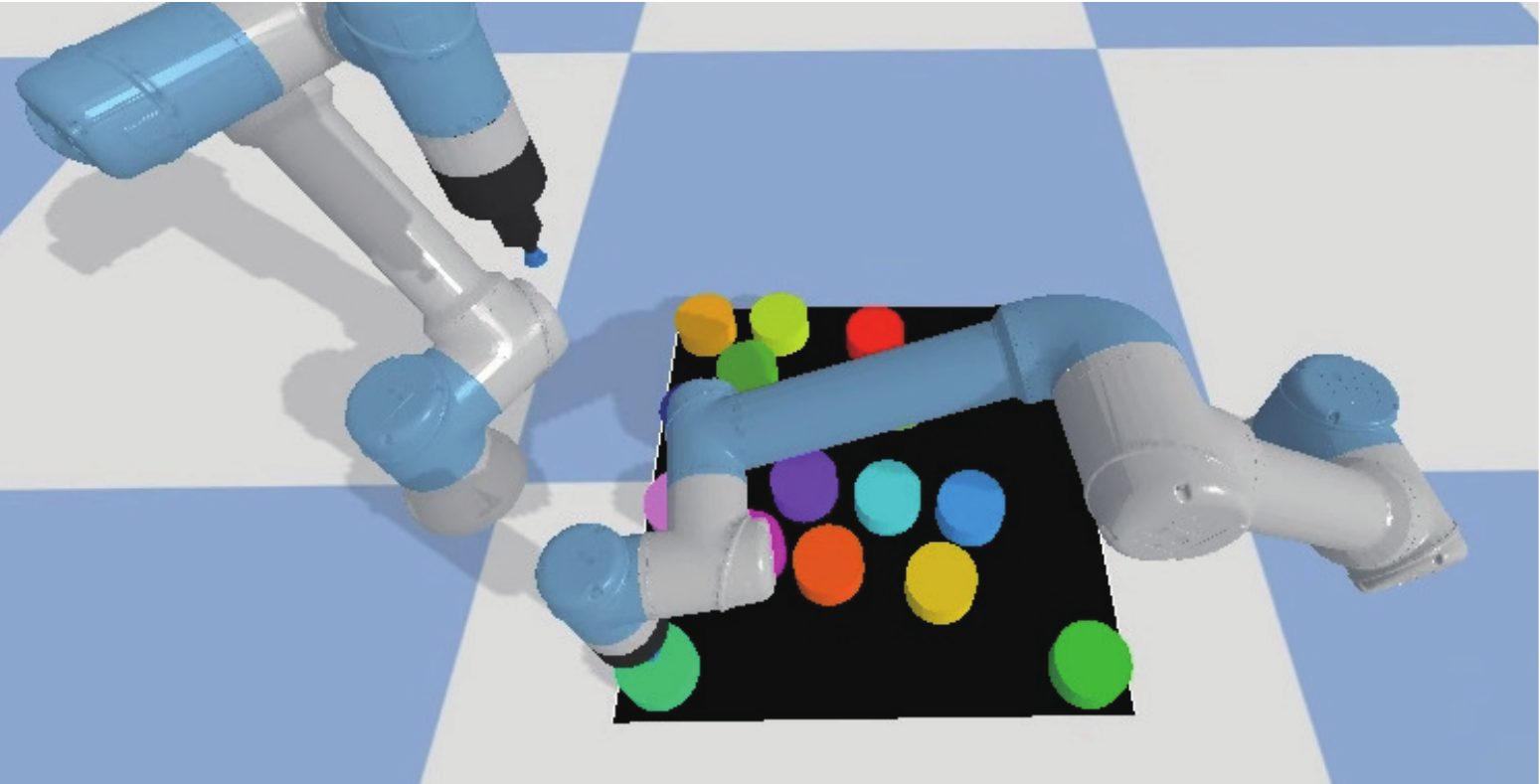}
    \caption{An instance of CDR with full overlap (CDRF), where each arm can reach every corner of the workspace but its manipulation is likely to be blocked by the other arm.}
    \label{fig:overlap_ratio}
\end{figure}

\begin{figure}[h]
    \vspace{-3mm}
    \centering
    \includegraphics[width=0.45\textwidth]{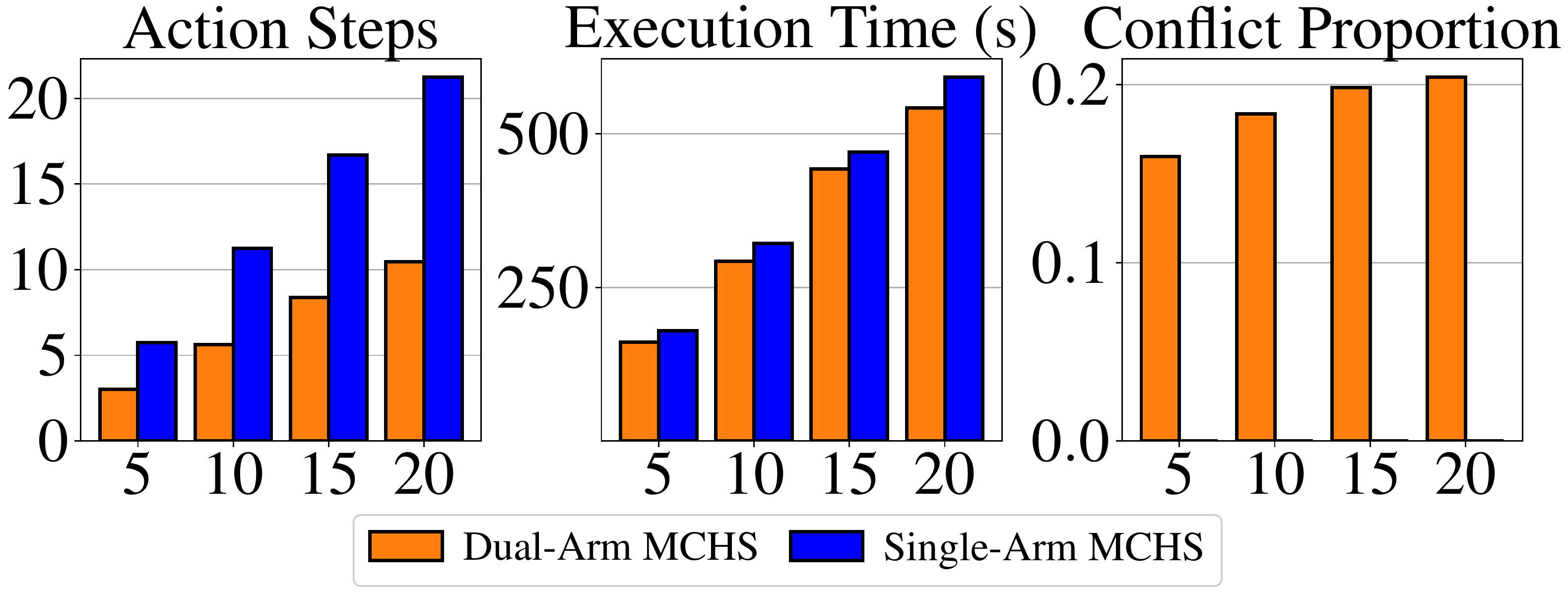}
    \caption{Algorithm performance in CDR instances with $\rho=1.0$. The x-axis represents the number of objects.}
    \label{fig:full_result}
    
\end{figure}

\textbf{Fully Overlapping Workspaces.} MCHS can also compute CDR plans with full overlap (CDRF), i.e., $\rho=1.0$.
As shown in Fig.~\ref{fig:overlap_ratio}, due to the reachability limit of UR-5e, the workspace is a bounded square with sides of length $0.6m$.
In this case, each arm can reach every corner of the workspace but its manipulation is likely to be blocked by the other arm.
We compare dual-arm MCHS plans to the single-arm MCHS plans, 
where $r_1$ take responsibility of all rearrangement tasks and the number of total actions is minimized.
The results (Fig.~\ref{fig:full_result}) indicate that each arm in the dual-arm system spends around $18\%$ of execution time yielding or making detours due to the blockage of the other arm.
Therefore, even though MCHS saves $50\%$ action steps, the execution of the plans is only around $10\%$ faster than that of the single-arm rearrangement plans.
However, the efficiency gain shown in action steps also suggests that the dual-arm system has the potential to save up to half of the execution time with specially designed arms for CDRF problems.

\vspace{-4mm}

\section{Conclusion}\label{sec:conclusion}
In this work, we investigate the cooperative dual-arm rearrangement (CDR) problem in a partially shared workspace.
We employ a lazy strategy for buffer allocation in the dual-arm rearrangement system.
To coordinate manipulations of the arms, we develop dependency graph guided heuristic search algorithms computing optimal primitive task schedules under two makespan evaluation metrics.
The effectiveness of the proposed procedure is demonstrated with extensive simulation studies in the PyBullet environment with two UR-5e robot arms.
Specifically, plans computed by our algorithms are up to $35\%$ shorter than greedy ones in execution time in dense rearrangement instances.
Moreover, we observe a tradeoff between the number of needed handoffs and the occurrence of path conflicts in CDR problems as the overlap ratio of the workspace varies.

{\small
\bibliographystyle{IEEEtran}
\bibliography{bib}
}

\section{Appendix}\label{sec:app}
\subsection{Interval State Space Heuristic Search for CDR in FC Makespan (FCHS)}\label{sec:CDR-KC}
FC metric estimates the total execution time of rearrangement.
For each pick-n-place action, the execution time is counted from the moment when the robot leaves for the object to the moment when the robot finishes the placement of the object.
Therefore, the cost of a pick-n-place can be expressed by $t_{pp}(p_c, p_g, p_r)$, 
where $p_c$ is the current end-effector pose. 
$p_g$, $p_r$ are the pick and place poses of the object.
Similarly, for each handoff operation, the execution time of the delivery (resp. receiving) arm is counted from the moment when the robot leaves for the pick pose (resp. handoff pose), 
to the moment when the handoff (resp. the placement) is finished.
With the same notations of locations, the cost of the delivery arm and the receiving arm in a handoff operation can be expressed as $t_{hd}(p_c, p_g)$ and $t_{hr}(p_c, p_r)$ respectively.
Besides object manipulations, robot arms move back to the rest pose when the rearrangement task is complete.

When the FC metric is considered, objects at intermediate states cannot be guaranteed to be at stable poses inside the workspace.
Based on the observation, we introduce a state space based on time intervals.
Each state can be expressed by two mappings $\mathcal L: \mathcal O \rightarrow \{S,G,\mathcal B(r_1), \mathcal B(r_2),T\}$ and $\mathcal T: \mathcal R\rightarrow \mathcal (O\times\{G,\mathcal B(r_1), \mathcal B(r_2)\}) \bigcup \{E\}$.
$\mathcal L$ indicates the current object status.
$S,G,\mathcal B(r_1), \mathcal B(r_2),T$ are five kinds of possible status of objects in the state: staying at start pose, goal pose, buffer in $\mathcal S(r_1)$, buffer in $\mathcal S(r_2)$, or being involved in the current task of a robot arm.
Specifically, $\mathcal L(o_i)=T$ when either $o_i$ is being manipulated by a robot, or a robot is on the way to pick it.
$\mathcal T$ indicates the current executing task of robot arms.
When $\mathcal T(r_i)=(o_j, G)$ (resp. $(o_j, \mathcal B(r_i))$), $r_i$ is executing a task moving $o_j$ from the current pose to its goal pose (resp. buffers).
When $\mathcal T(r_i)=E$, $r_i$ idles or just finishes an execution.
Each state except the start/goal state mimics a moment when one arm just placed an object while the other arm is executing a manipulation task or idling.

\begin{figure}[ht]
    \centering
    \includegraphics[width=0.48\textwidth]{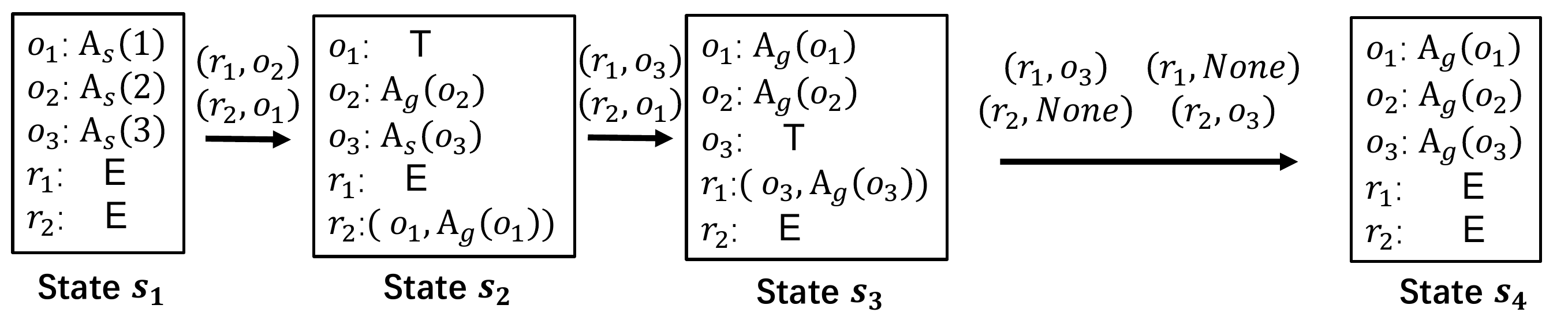}
    \includegraphics[width=0.2\textwidth]{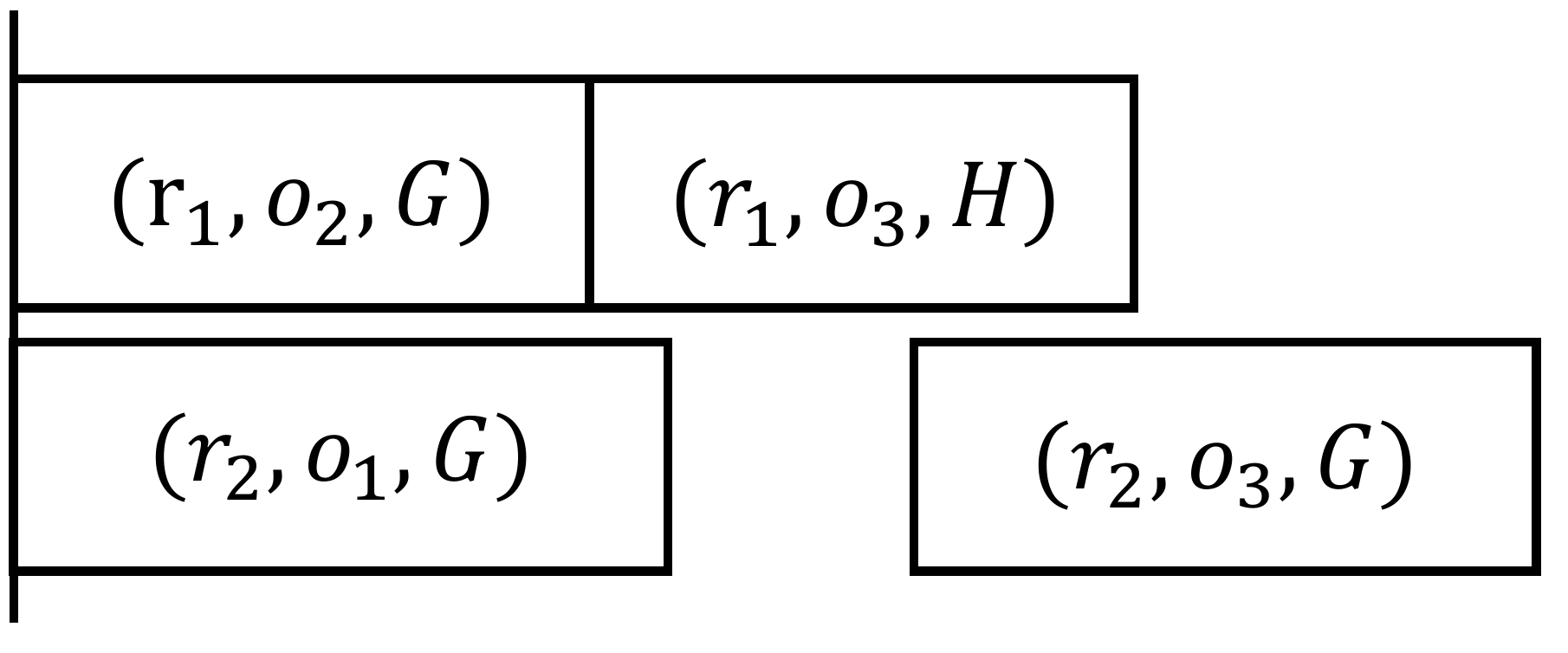}
    \caption{[Top] A path on the interval state space for the example instance in Fig.~\ref{fig:CDR-Exp}[Left]. [Bottom] The corresponding task schedule for the instance.}
    \label{fig:CDR-KC-Plan}
\end{figure}

A rearrangement plan of the instance in Fig.~\ref{fig:CDR-Exp}[Left], as a path in the interval state space, is presented in Fig.~\ref{fig:CDR-KC-Plan}.
The plan starts with the interval state $s_1$ representing the start arrangement.
In $s_1$, $r_1$ and $r_2$ choose to pick $o_2$ and $o_1$ respectively.
The second state in the path is the moment when $r_1$ completes the placement of $o_2$ while $r_2$ is still executing the manipulation.
When $r_2$ places $o_1$ at its goal pose, $r_1$ is attempting to move $o_3$ to the handoff pose, which yields to state $s_3$ in the path.
After the handoff, $r_2$ places $o_3$ at its goal pose and the plan ends when both arms return to their rest poses.

In either single arm rearrangement or dual-arm rearrangement with MC metric, object poses in arrangement states are determined, and the transition cost between states is the cost of specific manipulations. 
However, an interval state is more flexible and the transition cost between interval states is non-deterministic either. For example, in $s_3$ in Fig.~\ref{fig:CDR-KC-Plan}, without mentioning the path from the start state, $r_1$ can be on the way to $\mathcal A_s(o_3)$ or waiting at the handoff pose (since $\mathcal A_g(o_3)$ is out of reach). 
Moreover, there is no non-trivial lower bound on the transition cost between interval states. For example, from state $s_2$ to state $s_3$, the transition cost may be close to 0 if $r_2$ finishes the placement right after $r_1$ does. It may also be close to the cost of the pick-n-place if $r_2$ just starts to leave for $A_s(o_1)$ when $r_1$ finishes the placement.

Given a path on the search tree and an interval state $s$, the current cost $g(s)$ on the path is defined as the current makespan.
In the path presented in Fig.\ref{fig:CDR-KC-Plan}[Top], 
$g(s_2)=t_{pp}(\gamma_1, \mathcal A_s(o_2), \mathcal A_g(o_2))$,
where $\gamma_1$ is the projection of the rest pose of $r_1$'s end-effector to the 2d space embedding $\mathcal W$. It is the pick-n-place cost of $r_1$ to manipulate $o_2$. 
At this moment, $r_1$ completes the placement of $o_2$.
Similarly, $g(s_3)=t_{pp}(\gamma_2, \mathcal A_s(o_1), \mathcal A_g(o_1))$. 
At this moment, $r_2$ completes the placement of $o_1$.
Note that the buffer locations are treated as variables in this task scheduling phase. 
When a buffer is involved in a primitive action, e.g. moving an object from/to buffer, 
the travel distance of the arm is set as the diagonal distance of its reachable region.
This setting further penalizes temporary object displacements and ensures the computed schedule is executable no matter where the buffers are allocated in the workspace.
Based on the path from state $s_1$ to $s_4$, we obtain a corresponding task schedule.

Similar to the arrangement space heuristic search, for each interval state $s$, the heuristic function $h(s)$ is defined based on the cost of picking, transferring, placing, and handing off that are needed to rearrange objects which are still away from the goal poses. 
While the general idea is the same as manipulation cost based heuristic (Algo.~\ref{alg:MC-heuristic}), we make some modifications for interval states. 
The details are presented in Algo.\ref{alg:KC-heuristic}.
In an interval state, the ongoing task can be arbitrarily close to completion.
Therefore, besides the objects at the goal poses, we also ignore objects moving to the goal poses (Line 3),
When the object needs a handoff to the goal pose (Line 8), the delivering arm bears the delivery cost (Line 9-10), and the receiving arm bears the receiving cost (Line 11-12).

\begin{algorithm}
\begin{small}
    \SetKwInOut{Input}{Input}
    \SetKwInOut{Output}{Output}
    \SetKwComment{Comment}{\% }{}
    \caption{Full Cost Search Heuristic}
		\label{alg:KC-heuristic}
    \SetAlgoLined
		\vspace{0.5mm}
    \Input{$s$: current interval state}
    \Output{h: heuristic value of $s$}
		\vspace{0.5mm}
		cost[$r_1$], cost[$r_2$], sharedCost $\leftarrow 0,0,0$\\
		\For{$o_i\in \mathcal O$}{
		\lIf{$o_i$ is at goal or moving to goal}{continue}
		armSet $\leftarrow$ ReachableArms($s$, $o_i$, $\{\mathcal L(o_i), G\}$)\\
		\lIf{armSet is $\{r_1, r_2\}$}{sharedCost $\leftarrow$ sharedCost + transfer($o_i, \mathcal L(o_i), G$)+$t_g$+$t_r$}
		\lElseIf{armSet is $\{r_1\}$}{cost[$r_1$]$\leftarrow$ cost[$r_1$] + transfer($o_i, \mathcal L(o_i), G$)+$t_g$+$t_r$}
		\lElseIf{armSet is $\{r_2\}$}{cost[$r_2$]$\leftarrow$ cost[$r_2$] + transfer($o_i, \mathcal L(o_i), G$)+$t_g$+$t_r$}
		\Else{
		deliverArm $\leftarrow$ ReachableArms($s$, $o_i$, $\{\mathcal L(o_i)\}$)\\
		cost[deliverArm] $\leftarrow$ cost[deliverArm] + transfer($o_i, \mathcal L(o_i), H$)+$t_g$+$t_h$\\
		receiveArm $\leftarrow$ ReachableArms($s$, $o_i$, $\{\mathcal G\}$)\\
		cost[receiveArm] $\leftarrow$ cost[receiveArm] + transfer($o_i, H, G$)+$t_h$+$t_r$
		}
		}
		\vspace{1mm}
		\lIf{$\|$cost[$r_1$]-cost[$r_2$] $\|\leq$ sharedCost}{
		\\ \quad \Return (cost[$r_1$]+cost[$r_2$]+sharedCost)/2
		}
		\lElse{
		\Return max(cost[$r_1$], cost[$r_2$])
		}
\end{small}
\end{algorithm}



\end{document}